\documentclass[11pt,a4paper]{article}
\usepackage[hyperref]{emnlp2018}
\usepackage{times}
\usepackage{latexsym}
\usepackage{graphicx}

\usepackage{url}

\usepackage{enumitem}

\aclfinalcopy 

\usepackage{booktabs}
\usepackage{array}
\usepackage{adjustbox}
\usepackage[T1]{fontenc}
\usepackage{float}
\usepackage{algpseudocode}
\usepackage{algorithm}
\usepackage{caption}
\usepackage{amsmath,amssymb}
\DeclareMathOperator{\E}{\mathbb{E}}

\usepackage[colorinlistoftodos]{todonotes}
\renewcommand{\vec}[1]{\mathbf{#1}}
\usepackage{multirow}

\title{Adversarial Domain Adaptation for Duplicate Question Detection}

 \author{
  Darsh J Shah$^1$, Tao Lei$^2$, Alessandro Moschitti$^3$\thanks{\hspace{.3em} Work conducted while the author was at QCRI.},  \hspace{.1em}Salvatore Romeo$^4$, Preslav Nakov$^4$ \vspace{.2em}\\
    $^1$MIT CSAIL, Cambridge, MA, USA\\
    $^2$ASAPP Inc., New York, NY, USA \\
    $^3$Amazon, Manhattan Beach, CA, USA\\ 
    $^4$Qatar Computing Research Institute, HBKU, Doha, Qatar\\
{\tt darsh@csail.mit.edu, tao@asapp.com, amosch@amazon.com}\\
{\tt  sromeo@qf.org.qa, pnakov@qf.org.qa}}

\date{}

\begin{document}
\maketitle
\begin{abstract}
We address the problem of detecting duplicate questions in forums, which is an important step towards automating the process of answering new questions. As finding and annotating such potential duplicates manually is very tedious and costly, automatic methods based on machine learning are a viable alternative. However, many forums do not have annotated data, i.e.,~questions labeled by experts as duplicates, and thus a promising solution is to use domain adaptation from another forum that has such annotations.
Here we focus on adversarial domain adaptation, deriving important findings about when it performs well and what properties of the domains are important in this regard. 
Our experiments with $\emph{StackExchange}$ data show an average improvement of 5.6\% over the best baseline across multiple pairs of domains.
\end{abstract}

\section{Introduction}

Recent years have seen the rise of community question answering forums, which allow users to ask questions and to get answers 
in a collaborative fashion. One issue with such forums is that duplicate questions easily become ubiquitous as users often ask the same question, possibly in a slightly different formulation, making it difficult to find the best (or one correct) answer \cite{cQA:Survey:2018,lai-bui-li:2018:C18-1}. Many forums allow users to signal such duplicates, but this can only be done after the duplicate question has already been posted and has possibly received some answers, which complicates merging the question threads. 
Discovering possible duplicates at the time of posting is much more valuable from the perspective of both (\emph{i})~the forum, as it could prevent a duplicate from being posted,
and (\emph{ii})~the users, as they could get an answer immediately.

\noindent Duplicate question detection is a special case of the more general problem of question-question similarity. The latter was addressed using a variety of textual similarity measures, topic modeling \cite{Cao:2008:RQU:1367497.1367509,zhang2014question},
and syntactic structure \cite{wang2009syntactic,SemEval2016:task3:KeLP,DaSanMartino:CIKM:2016,SemEval2016:task3:ConvKN,SemEval-2017:task3:KELP}. Another approach is to use neural networks such as feed-forward \cite{nakov-marquez-guzman:2016:EMNLP2016}, convolutional \cite{dossantos-EtAl:2015:ACL-IJCNLP,bonadiman-uva-moschitti:2017:EACLshort,Wang:2018:CAC:3183892.3151957}, long short-term memory \cite{Romeo:2016coling}, and more complex models \cite{LeiJBJTMM16,DBLP:conf/cikm/NicosiaM17,P18-2046,Joty:2018:multitask,Zhang:Wu:2018}.
Translation models have also been popular 
\cite{zhou2011phrase,Jeon:2005:FSQ:1099554.1099572,ACL2016:MTE-NN-cQA,SemEval2016:task3:MTE-NN}.

The above work assumes labeled training data, which exists for question-question similarity, e.g.,~from SemEval-2016/2017 \cite{agirre-EtAl:2016:SemEval1,nakov-EtAl:2016:SemEval,SemEval-2017:task3}, and for duplicate question detection, e.g.,~SemEval-2017 task 3 featured four StackExchange forums, \emph{Android}, \emph{English}, \emph{Gaming}, and \emph{Wordpress}, from 
CQADupStack \cite{hoogeveen2015cqadupstack,Hoogeveen+:2016a}. Yet, such annotation is not available for many other forums, e.g.,~the \emph{Apple} community on StackExchange. 

In this paper, we address this lack of annotation using adversarial domain adaptation (ADA) to effectively use labeled data from another forum. 
Our contributions can be summarized as follows:
\vspace{-3pt}
\begin{itemize}[leftmargin=10pt]
  \setlength\itemsep{-5pt}
  \item we are the first to apply adversarial domain adaptation to the problem of duplicate question detection across different domains;\footnote{The code and the data are available at the following link:\\ \url{http://github.com/darsh10/qra_code}}
  \item on the StackExchange family of forums, our model outperforms the best baseline with an average relative improvement of 5.6\%  (up to 14\%) across all domain pairs.
  \item we study when transfer learning performs well and what properties of the domains are important in this regard; and 
  \item we show that adversarial domain adaptation can be efficient even for unseen target domains, given some similarity of the target domain with the source one and with the regularizing adversarial domain.
\end{itemize}

Adversarial domain adaptation (ADA) was proposed by \citet{ganin2015unsupervised}, and was then used for NLP tasks such as sentiment analysis and retrieval-based question answering \cite{chen2016adversarial,Ganin:2016:DTN:2946645.2946704,Li:2017:EAM:3172077.3172199,liu2017adversarial,yu2017modelling,zhang2017aspect}, including cross-language adaptation \cite{joty-EtAl:2017:CoNLL} for question-question similarity.\footnote{Prior work on cross-language adaptation for question-question similarity used cross-language tree kernels \cite{DaSanMartino:2017:CQR:3077136.3080743}.}

The rest of this paper is organized as follows: Section 2 presents our model, its components, and the training procedure. Section 3 describes the datasets we used for our experiments, stressing upon their nature and diversity. Section 4 describes our adaptation experiments and discusses the results. Finally, Section 5 concludes with possible directions for future work.
\section{Method}
\label{qsnn}

\begin{figure}[t]
\center
  \includegraphics[width=1.0\linewidth]{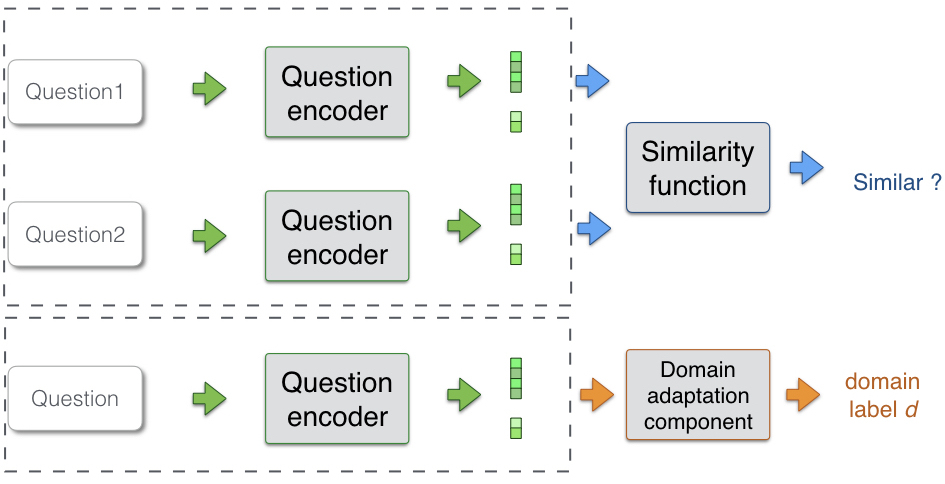}
  \caption{\small Our cross-domain question-question similarity model. The Question encoder is common for the questions from the source domain and from the target domain. The model and its training procedure are described in Section 2.}
  \label{fig:model}
\end{figure}

Our ADA model has three components: (\emph{i})~question encoder, (\emph{ii})~similarity function, and (\emph{iii})~domain adaptation component, as shown in Figure~\ref{fig:model}. 

The encoder $E$ maps a sequence of word tokens $x=(x_{1},..,x_{n})$
to a dense
vector $\vec v = E(x)$. 
The similarity function $f$ takes two question vectors, $\vec v_1$ and $\vec v_2$, and predicts whether the corresponding questions are duplicates. 

\noindent The domain classifier $g$ takes a question vector $\vec v$ and predicts whether the question is from the source or from the target domain.
We train the encoder not only to do well on the task for the source data, but also to fool the domain classifier, as shown in Algorithm~1. 
We describe the design choices considered for our domain adaptation model in the following two subsections.

\subsection{Question Similarity Function}

\label{simmodel}
We consider two options for our similarity function $f(\vec v_{1}, \vec v_{2})$:
\newline
\noindent{(\emph{i})}~a logistic function that computes the probability that two questions are similar/duplicates, which is trained with the cross-entropy loss:
\vspace{-0.5em}
\begin{align*}
\text{sigmoid}\left({\vec W}^\top ({\vec v_1}\odot{\vec v_2}) + {\vec b}\right)
\end{align*}
where $\odot$ is an element-wise vector product between unit encodings of questions; \newline

\noindent (\emph{ii)}~a simple cosine similarity function, i.e.,~$cosine( \vec v_{1}, \vec v_{2})$, trained using the pairwise hinge loss with a margin $m$:
\vspace{-0.5em}
$$\sum_{i} max(\{(1-y^{i})f(\vec v^{i}_{1}, \vec v^{i}_{2}) + m -  y^{i}f(\vec v^{i}_{1}, \vec v^{i}_{2})\} ,0)$$

Our experiments reported in Table~\ref{table:2} show that the $cosine$ similarity function performs far better. 

\subsection{Domain Adaptation Components}

The adversarial component is responsible for reducing the difference between the source and the target domain distributions. There are two common approaches to achieve this:  (\emph{i})~classification-based~\cite{ganin2015unsupervised} and  (\emph{ii})~Wasserstein~\cite{arjovsky2017wasserstein}.

The main difference between them is in the way the domain discrepancy loss is computed.
In the classification-based approach, the adversarial component is a classifier trained to correctly predict the domain (source vs.~target) of the input question. In contrast, the question encoder is optimized to confuse the domain classifier, which, as a result, encourages domain invariance. \newcite{arjovsky2017towards} showed that this adversarial optimization process resembles minimizing the Jenson-Shannon (JS) divergence between the source $P_{s}$ and the target distribution $P_{t}$: 
$$JS(P_{s}, P_{t}) = KL(P_{s},P_{m}) + KL(P_{g},P_{m})$$

\noindent where $P_{m} = (P_{s}+P_{t})/2 $ and KL is the Kullback-Leibler divergence.

\noindent In contrast, the Wasserstein method attempts to reduce the approximated Wasserstein distance (also known as \emph{Earth Mover's Distance}) between the distributions for the source and for the target domain as follows:
$$W(P_{s}, P_{t}) = \underset{||f||_{L}\leq1}{sup} \E_{x\sim{}P_s}[f(x)] - \E_{x\sim{}P_t}[f(x)]$$

\noindent where $f$ is a Lipchitz-1 continuous function realized by a neural network.

\citet{arjovsky2017wasserstein} have shown that the Wasserstein method yields more stable training for computer vision tasks.

\subsection{Training Procedure}
Algorithm 1 describes the procedure to train the three components of our model. Adversarial training needs two kinds of training data: (\emph{i})~annotated question pairs from the source domain, and (\emph{ii})~unlabeled questions from the source and the target domains. 

The question encoder is trained to perform well on the source domain using the similarity classification loss $L_{c}$. In order to enforce good performance on the target domain, the question encoder is simultaneously trained to be incapable in discriminating between question pairs from the source vs. the target domain. This is done through the domain classification loss $L_{d}$.
\begin{algorithm}[t]
\label{alg:training}
\footnotesize
\caption*{\textbf{Algorithm 1}: Training Procedure }
\begin{algorithmic}
\State Input: source data $X^{s}$; target data $X^{t}$ 
\State Hyper-parameters: learning rates $\alpha_{1},\alpha_{2}$; batch size $m$; adversarial importance $\lambda$ 
\State Parameters to be trained: question encoder $\theta_{e}$, question similarity classifier $\theta_{s}$ and domain classifier $\theta_{d}$
\State Similarity classification loss $L_{c}$ is either the cross-entropy loss or hinge loss, described in Section 2.1
\State Adversarial loss $L_{d}$, described in Section 2.2
\Repeat
\For{\texttt{each batch}}
\State Construct a sub-batch of similar and dissimilar question pairs from the annotated source data $\{(x_{i_1}^s,x_{i_2}^s),y_i^s\}_{i=1}^m$
\State Calculate the classification loss $L_{c}$ using $\theta_{e}$ and $\theta_{s}$ for this sub-batch
\State Construct a sub-batch of questions $\{x_{i}^s,x_{j}^t\}_{i=1}^m$ from the corpora of source and target domains
\State Calculate the domain discrepancy loss $L_{d}$ using $\theta_{e}$ and $\theta_{d}$ for this sub-batch
\State Total loss $L = L_{c} - \lambda L_{d}$
\State $\theta_{e} = \theta_{e} - \alpha_{1} {\triangledown_{\theta_{e}}{L}}$
\State $\theta_{s} = \theta_{s} - \alpha_{1} {\triangledown_{\theta_{s}}{L}}$
\State $\theta_{d} = \theta_{d} + \alpha_{2} {\triangledown_{\theta_{d}}{L}}$
\EndFor
\Until{ $\theta_{e}$,  $\theta_{s}$ and  $\theta_{d}$ converge}
\end{algorithmic}
\end{algorithm}

\begin{table}[t]
\begin{center}
\begin{adjustbox}{max width=\linewidth}
\begin{tabular}{ m{4.5em} m{3.9em} m{4em} m{2.5em} m{2.5em} m{2.5em} } 
\toprule
\bf Dataset & \bf Questions & \bf Duplicates & \bf Train & \bf Dev & \bf Test\\ 
\midrule
AskUbuntu & 257,173 & 27,289 & 9,106  & 1,000  & 1,000 \\ 
SuperUser & 343,033 & 11,407 & 9,106  & 1,000  & 1,000  \\
Apple & 80,466 & 2,267 & -- & 1,000  & 1,000  \\
Android & 42,970 & 2,371 & -- & 1,000  & 1,000  \\
\midrule
Sprint & 31,768 & 23,826 & 9,100 & 1,000& 1,000 \\
\midrule
Quora & 537,211 & 149,306 & 9,100 & -- & -- \\
\bottomrule
\end{tabular}
\end{adjustbox}
\end{center}

\caption{\small Statistics about the datasets. The table shows the number of question pairs that have been manually marked as similar/duplicates by the forum users (i.e.,~positive pairs). We further add 100 negative question pairs per duplicate question by randomly sampling from the full corpus of questions.}
\label{table:6}
\end{table}

\section{Datasets}

The datasets we use can be grouped as follows:

\begin{itemize}[leftmargin=10pt]
\item{\bf Stack Exchange} is a family of technical community support forums. We collect questions (composed by title and body) from the XML dumps of four forums: \emph{AskUbuntu}, \emph{SuperUser}, \emph{Apple}, and \emph{Android}. Some pairs of similar/duplicate questions in these forums are marked by community users.

\item{\bf Sprint FAQ} is a newly crawled dataset from the Sprint technical forum website. It contains a set of frequently asked questions and their paraphrases, i.e., three similar questions, paraphrased by annotators.

\item{\bf Quora} is a dataset of pairs of similar questions asked by people on the Quora website. They cover a broad set of topics touching upon philosophy, entertainment and politics.
\end{itemize}

Note that these datasets are quite heterogeneous: the \emph{StackExchange} forums focus on specific technologies, where questions are informal and users tend to ramble on about their issues, the \emph{Sprint FAQ} forum is technical, but its questions are concise and shorter, and the \emph{Quora} forum covers many different topics, including non-technical. 

Statistics about the datasets are shown in Table~\ref{table:6}. Moreover, in order to quantify the differences and the similarities, we calculated the fraction of unigrams, bigrams and trigrams that are shared by pairs of domains.
Table~\ref{table:0} shows statistics about the $n$-gram overlap between \emph{AskUbuntu} or \emph{Quora} as the source and all other domains as the target. As one might expect, there is a larger overlap within the \emph{StackExchange} family.

\section{Experiments and Evaluation}

\subsection{Experimental Setup}

\begin{table}[t]
\begin{center}
\footnotesize
\begin{adjustbox}{max width=1.05\linewidth}
\begin{tabular}{  m{4.2em}  m{4.2em}  m{3.4em}  m{3.2em}  m{3.2em} } 
\toprule
\multicolumn{1}{c}{\bf Source}  & \multicolumn{1}{c}{\bf Target}  & \multicolumn{1}{c}{\bf Unigrams} & \multicolumn{1}{c}{\bf Bigrams}  & \multicolumn{1}{c}{\bf Trigrams} \\ 
\hline
\multirow{4}{*}{AskUbuntu} & Android &0.989 & 0.926 & 0.842\\ 

           & Apple & 0.991&0.926&0.853\\ 

           & SuperUser & 0.990&0.921&0.822\\ 
\cline{2-5}
          & Sprint &0.959  &0.724&0.407\\
           
          & Quora & 0.922  &0.696  &0.488\\

\hline
\multirow{4}{*}{Quora}  & AskUbuntu & 0.949&0.647&0.326\\ 

          & Apple & 0.969&0.721&0.426\\ 

          & Android & 0.973&0.762&0.473\\ 
          & SuperUser & 0.958 & 0.663 & 0.338\\ 
\cline{2-5}
                          & Sprint & 0.942&0.647 & 0.310 \\
\bottomrule
\end{tabular}
\end{adjustbox}
\end{center}
\caption{\footnotesize Proportion of $n$-grams that are shared between the source and the target domains.}
\label{table:0}
\end{table}

\textbf{Baselines}
We compare our ADA model to the following baselines:
(\emph{a})~\emph{direct transfer}, which directly applies models learned from the source to the target domain without any adaptation; and (\emph{b})~the standard  unsupervised \emph{BM25} ~\cite{robertson2009probabilistic} scoring provided in search engines such as Apache Lucene ~\cite{mccandless2010lucene}. 

\textbf{Models}
We use a bi-LSTM~\cite{hochreiter1997long} encoder that operates on 300-dimensional GloVe word embeddings ~\cite{pennington2014glove}, which we train on the combined data from all domains. We keep word embeddings fixed in our experiments. For the adversarial component, we use a multi-layer perceptron.

\textbf{Evaluation Metrics}
As our datasets may contain some duplicate question pairs, which were not discovered and thus not annotated, we end up having false negatives. Metrics such as MAP and MRR are not suitable in this situation.
Instead, we use AUC (area under the curve) to evaluate how well the model ranks positive pairs vs. negative ones. AUC quantifies how well the true positive rate ($tpr$) grows at various false positive rates ($fpr$) by calculating the area under the curve starting from $fpr = 0$ to $fpr = 1$.
We compute the area integrating the false positive rate ($x$-axis) from $0$ up to a threshold $t$, and we normalize the area to $[0,1]$. This score is known as AUC$(t)$.
It is more stable than MRR and MAP in our case when there could be several false negatives.\footnote{For illustration, say of 100 candidates, 2 false negatives are ranked higher than the correct pair, the AUC score drops by 3 points (linear drop), as compared to the 66.67 point drop for MRR. We can avoid the expensive manually tagging of the negative pairs for experiments by using the AUC score.}

\begin{table}[t]
\begin{center}
\footnotesize
\begin{tabular}{cccc} 
\toprule
\bf Adaptation       & \bf Similarity  &  \bf AUC(0.05) & \bf AUC(0.1)\\ 
\hline
--- & Sigmoid   & 0.431  & 0.557 \\ 
--- & Cosine    & 0.692 & 0.782 \\ 
\hline
Classification & Cosine  &  0.791 & 0.862 \\ 
Wasserstein & Cosine & 0.795 & 0.869 \\ 
\bottomrule
\end{tabular}
\end{center}
\vspace{-.5em}
\caption{\footnotesize Duplicate question detection: direct transfer vs. adversarial domain adaptation from \emph{AskUbuntu} to \emph{Android}.
}
\label{table:2}
\end{table}

\subsection{Choosing the Model Components}

\textbf{Model Selection}
We select the best components for our domain adaptation model via experimentation on the \emph{AskUbuntu}--\emph{Android} domain pair. Then, we apply the model with the best-performing components across all domain pairs. 

\textbf{Hyperparameters}
We fine-tune the hyper-parameters of all models on the development set for the target domain.

\textbf{Similarity Function}
Table \ref{table:2} shows the AUC at 0.05 and 0.1 for different models of question similarity, training on \emph{AskUbuntu} and testing on \emph{Android}. The first row shows that using cosine similarity with a hinge loss yields much better results than using a cross-entropy loss. This is likely because (\emph{i})~there are some duplicate question pairs that were not tagged as such and that have come up as negative pairs in our training set, and the hinge loss deals with such outliers better. (\emph{ii})~The cosine similarity is domain-invariant, while the weights of the feed-forward network of the softmax layers capture source-domain features.

\textbf{Domain Adaptation Component}
We can see that the Wasserstein and the classification-based methods perform very similarly, after proper hyper-parameter tuning. However, Wasserstein yields better stability, achieving an AUC variance 17 times lower than the one for classification across hyper-parameter settings. Thus, we chose it for all experiments in the following subsections.

\begin{table}[t]
\begin{center}
\footnotesize
\begin{tabular}{  m{4.5em}  m{4.5em}  m{2.6em}  m{2.5em}  m{2.5em} } 
\toprule
\bf Source  & \bf Target  & \bf Direct & \bf BM25  & \bf Adv. \\ 
\hline
\multirow{4}{*}{AskUbuntu}  & Android & 0.692 & 0.681 & \textbf{0.790}\\ 
            & Apple & 0.828 & 0.747 & \textbf{0.855}\\ 
            & SuperUser & 0.908 & 0.765 & \textbf{0.911}\\ 
                            & Sprint & 0.917 & \textbf{0.956} & 0.937 \\ 
\hline
\multirow{3}{*}{SuperUser}  & AskUbuntu & 0.730 & 0.644 & \textbf{0.796}\\ 
            & Apple & 0.828 & 0.747 & \textbf{0.861}\\ 
            & Android & 0.770 & 0.681 & \textbf{0.790}\\ 
                            & Sprint & 0.928 & \textbf{0.956} & 0.932 \\
\bottomrule
\end{tabular}
\end{center}
\vspace{-.5em}
\caption{\footnotesize Domain adaptation for the \emph{StackExchange} source-target domain pairs when using the \emph{Direct} approach, \emph{BM25}, and our adaptation model, measured with AUC(0.05).}
\label{table:3}
\end{table}

\subsection{When Does Adaptation Work Well?}

Tables~\ref{table:3} and ~\ref{table:4} study the impact of domain adaptation when applied to various source-target domain pairs,  using the \emph{Direct} approach, \emph{BM25}, and our adaptation model. 
We can make the following observations: 

\begin{itemize}
\item For almost all source--target domain pairs from the \emph{StackExchange} family, domain adaptation improves over both baselines, with an average relative improvement of 5.6\%. This improvement goes up to 14\% for the \emph{AskUbuntu}--\emph{Android} source--target domain pair.

\item Domain adaptation on the \emph{Sprint} dataset performs better than direct transfer, but it is still worse than \emph{BM25}.

\item Domain adaptation from \emph{Quora} performs the worst, with almost no improvement over direct transfer, which is far behind \emph{BM25}. 

\item The more similar the source and the target domains, the better our adaptation model performs.
\end{itemize}

Table~\ref{table:0} shows that \emph{AskUbuntu} has high similarity to other \emph{StackExchange} domains, lower similarity to \emph{Sprint}, and even lower similarity to \emph{Quora}. The Pearson coefficient \cite{myers2010research} between the $n$-gram fractions and the domain adaptation effectiveness for unigrams, bigrams and trigrams is 0.57, 0.80 and 0.94, respectively, which corresponds to moderate-to-strong positive correlation. This gives insight into how simple statistics can predict the overall effectiveness of domain adaptation.

\begin{table}[t]
\begin{center}
\footnotesize
\begin{tabular}{  m{4.5em}  m{4.5em}  m{2.6em}  m{2.5em}  m{2.5em} } 
\toprule
\bf Source  & \bf Target  & \bf Direct & \bf BM25  & \bf Adv. \\ 
\hline
\multirow{3}{*}{Sprint}     & AskUbuntu   & 0.615 & \textbf{0.644} & 0.615\\ 
            & Apple & 0.719 & \textbf{0.747} & 0.728\\ 
            & Android & 0.627 & \textbf{0.681} & 0.648\\ 
                            & Sprint & 0.977 & 0.956 & -- \\
            & SuperUser & 0.795 & 0.765 & \textbf{0.795}\\ 
\hline
\multirow{3}{*}{Quora}    & AskUbuntu & 0.446 & \textbf{0.644} & 0.446 \\ 
            & Apple & 0.543 & \textbf{0.747} & 0.543 \\ 
            & Android & 0.443 & \textbf{0.681} & 0.460 \\ 
                            & Sprint & 0.786 & \textbf{0.956} & 0.794 \\
            & SuperUser & 0.624 & \textbf{0.765} & 0.649 \\ 
\bottomrule
\end{tabular}
\end{center}
\vspace{-.5em}
\caption{\footnotesize Domain adaptation results when using \emph{Sprint} and \emph{Quora} as the source domains with the \emph{Direct} approach, \emph{BM25}, and our adaptation model, measured with AUC(0.05).}
\label{table:4}
\end{table}

\begin{table}[h]
\begin{center}
\footnotesize
\begin{tabular}{  m{5em}  m{4.5em}  m{2.5em} m{3.5em}  } 
\toprule
{\textbf{Pivot}}$\backslash${\textbf{Target}}  & SuperUser & Apple  & Android \\ 
\hline
SuperUser   & 0.911 & 0.827 & 0.678 \\ 
Apple   & 0.900 & 0.855 & \textbf{0.711} \\ 
Android     & 0.904 & \textbf{0.843} & 0.790 \\ 
Quora     & 0.906 & 0.815 & 0.673 \\ 
\hline
Direct    & 0.908 & 0.828 & 0.692 \\
\bottomrule
\end{tabular}
\end{center}
\vspace{-.5em}
\caption{\footnotesize AUC(0.05) of ADA to unseen domains, with \emph{AskUbuntu} as a source.}
\label{table:7}
\end{table}

\subsection{Adapting to Unseen Domains}

We also experiment with domain adaptation to a target domain that was not seen during training (even adversarial training). We do so by training to adapt to a pivot domain different from the target. Table~\ref{table:7} shows that this yields better AUC compared to \emph{direct transfer} when using \emph{Apple} and \emph{Android} as the pivot/target domains. 
We hypothesize that this is due to \emph{Apple} and \emph{Android} being closely related technical forums for iOS and Android devices.
This sheds some light on the generality of adversarial regularization.

\section{Conclusion and Future Work} 

We have applied and analyzed adversarial methods for domain transfer for the task of duplicate question detection; to the best of our knowledge, this is the first such work.
Our experiments suggest that (\emph{i})~adversarial adaptation is rather effective between domains that are similar, and (\emph{ii})~the effectiveness of adaptation is positively correlated with the $n$-gram similarity between the domains. 

In future work, we plan to develop better methods for adversarial adaptation based on these observations. One idea is to try source-pivot-target transfer, similarly to the way this is done for machine translation \cite{Wu2007}. 
Another promising direction is to have an attention mechanism \cite{luong2015effective} for question similarity which can be adapted across domains.\footnote{In our experiments, we found that using attention was lowering the adaptation performance. Upon analysis, we found that adversarial techniques alone were not enough to make the attention weights domain-invariant.}

\section*{Acknowledgments}
This research was carried out in collaboration between the MIT Computer Science and Artificial Intelligence Laboratory (CSAIL) and the Qatar Computing Research Institute (QCRI), HBKU.

\bibliography{bibliography}
\bibliographystyle{acl_natbib_nourl}

\end{document}